\documentclass[runningheads]{llncs}

\usepackage{graphicx}
\usepackage{url}
\usepackage{color,xcolor}

\usepackage{amsmath}
\usepackage{esvect}
\usepackage{bm}

\begin{document}

\title{Predicting Fluid Intelligence of Children using T1-weighted MR Images and a StackNet}
\titlerunning{Predicting Gf of Children using MR-T1 images and StackNet}
\author{Po-Yu Kao\inst{1}\orcidID{0000-0002-9439-8819} \and
Angela Zhang\inst{1} \and
Michael Goebel\inst{1} \and \protect\\
Jefferson W. Chen\inst{2} \and 
B.S. Manjunath\inst{1}}

\authorrunning{P.-Y. Kao et al.}

\institute{University of California, Santa Barbara, California, United States \\ \email{\{poyu\_kao,manj\}@ucsb.edu} \and
University of California, Irvine, California, United States
}

\maketitle
\setcounter{footnote}{0}

\begin{abstract}
In this work, we utilize T1-weighted MR images and StackNet to predict fluid intelligence in adolescents. Our framework includes feature extraction, feature normalization, feature denoising, feature selection, training a StackNet, and predicting fluid intelligence. The extracted feature is the distribution of different brain tissues in different brain parcellation regions. The proposed StackNet consists of three layers and 11 models. Each layer uses the predictions from all previous layers including the input layer. The proposed StackNet is tested on a public benchmark Adolescent Brain Cognitive Development Neurocognitive Prediction Challenge 2019 and achieves a mean squared error of 82.42 on the combined training and validation set with 10-fold cross-validation. The proposed StackNet achieves a mean squared error of 94.25 on the testing data. The source code is available on GitHub\footnote{\url{https://github.com/UCSB-VRL/ABCD-MICCAI2019}}.

\end{abstract}

\section{Introduction}

Fluid intelligence (Gf) refers to the ability to reason and to solve new problems independently of previously acquired knowledge. Gf is critical for a wide variety of cognitive tasks, and it is considered one of the most important factors in learning. Moreover, Gf is closely related to professional and educational success, especially in complex and demanding environments \cite{jaeggi2008improving}. The ABCD Neurocognitive Prediction Challenge (ABCD-NP-Challenge 2019) provides 8556 subjects, age 9-10 years, with T1-weighted MR images and fluid intelligence which is withheld for testing subjects. The motivation of the ABCD-NP-Challenge 2019 is to discover the relationship between the brain and behavioral measures by leveraging the modern machine learning methods. 

A few recent studies use structural MR images to predict fluid intelligence. Paul et al. \cite{paul2016dissociable} demonstrated that brain volume is correlated with quantitative reasoning and working memory. Wang et al. \cite{wang2015mri} proposed a novel framework for the estimation of a subject's intelligence quotient score with sparse learning based on the neuroimaging features. In this work, we utilize the T1-weighted MR images of adolescents to predict their fluid intelligence with a StackNet. While whole brain volumes have been examined in relation to aspects of intelligence, to our knowledge there has been no previous work which examines the predictive ability of whole brain parcellation distributions for fluid intelligence. The main contributions of our work are two-fold: (1) to predict pre-residualized fluid intelligence based on parcellation volume distributions, and (2) to show the significance of the volume of each region on the overall prediction.

\section{Materials and Methods}

\subsection{Dataset}

The Adolescent Brain Cognitive Development Neurocognitive Prediction Challenge (ABCD-NP-Challenge 2019) \cite{garavan2018recruiting,hagler2018image,luciana2018adolescent,pfefferbaum2017altered,volkow2018conception} provides data for 3739 training subjects, 415 validation subjects and 4402 testing subjects (age 9-10 years). MR-T1 image is given for each subject, but the fluid intelligence scores are only provided for the training and validation subjects. MR-T1 images are distributed after skull-stripped and registered to the SRI 24 atlas \cite{rohlfing2010sri24} of voxel dimension $240 \times 240 \times 240$. In addition to the MR-T1 images, the distributions of gray matter, white matter, and cerebrospinal fluid in different regions of interest according to the SRI 24 atlas are also provided for all subjects. The fluid intelligence scores are pre-residualized on data collection site, sociodemographic variables and brain volume. The provided scores should, therefore, represent differences in Gf not due to these known factors. 

Data used in the preparation of this article were obtained from the Adolescent Brain Cognitive Development (ABCD) Study (\url{https://abcdstudy.org}), held in the NIMH Data Archive (NDA). 
This is a multisite, longitudinal study designed to recruit more than 10,000 children age 9-10 and follow them over 10 years into early adulthood.
The ABCD Study is supported by the National Institutes of Health and additional federal partners under award numbers U01DA041022, U01DA041028, U01DA041048, U01DA041089, U01DA041106, U01DA041117, U01DA041120, U01DA041134, U01DA041148, U01DA041156, U01DA041174, U24DA041123, U24DA041147, U01DA041093, and U01DA041025. 
A full list of supporters is available at \url{https://abcdstudy.org/federal-partners.html}. 
A listing of participating sites and a complete listing of the study investigators can be found at \url{https://abcdstudy.org/Consortium_Members.pdf}. 
ABCD consortium investigators designed and implemented the study and/or provided data but did not necessarily participate in analysis or writing of this report. 
This manuscript reflects the views of the authors and may not reflect the opinions or views of the NIH or ABCD consortium investigators.
The ABCD data repository grows and changes over time. 
The ABCD data used in this report came from DOI:10.15154/1518671. DOIs can be found at \url{http://dx.doi.org/10.15154/1518671}.

\subsection{StackNet Design}

StackNet \cite{michailidis2017stacknet} is a computational, scalable and analytical framework that resembles a feed-forward neural network. It uses Wolpert's stacked generalization \cite{wolpert1992stacked} in multiple levels to improve the accuracy of classifier or reduce the error of regressor. In contrast to the backward propagation used by feed-forward neural networks during the training phase, StackNet is built iteratively one layer at a time (using stacked generalization), with each layer using the final target as its target. 

There are two different modes of StackNet: (i) each layer directly uses the predictions from only one previous layer, and (ii) each layer uses the predictions from all previous layers including the input layer that is called restacking mode. StackNet is usually better than the best single model contained in each first layer. However, its ability to perform well still relies on a mix of strong and diverse single models in order to get the best out of this meta-modeling methodology.

\begin{figure}[htbp]
    \centering
    \includegraphics[width=\linewidth]{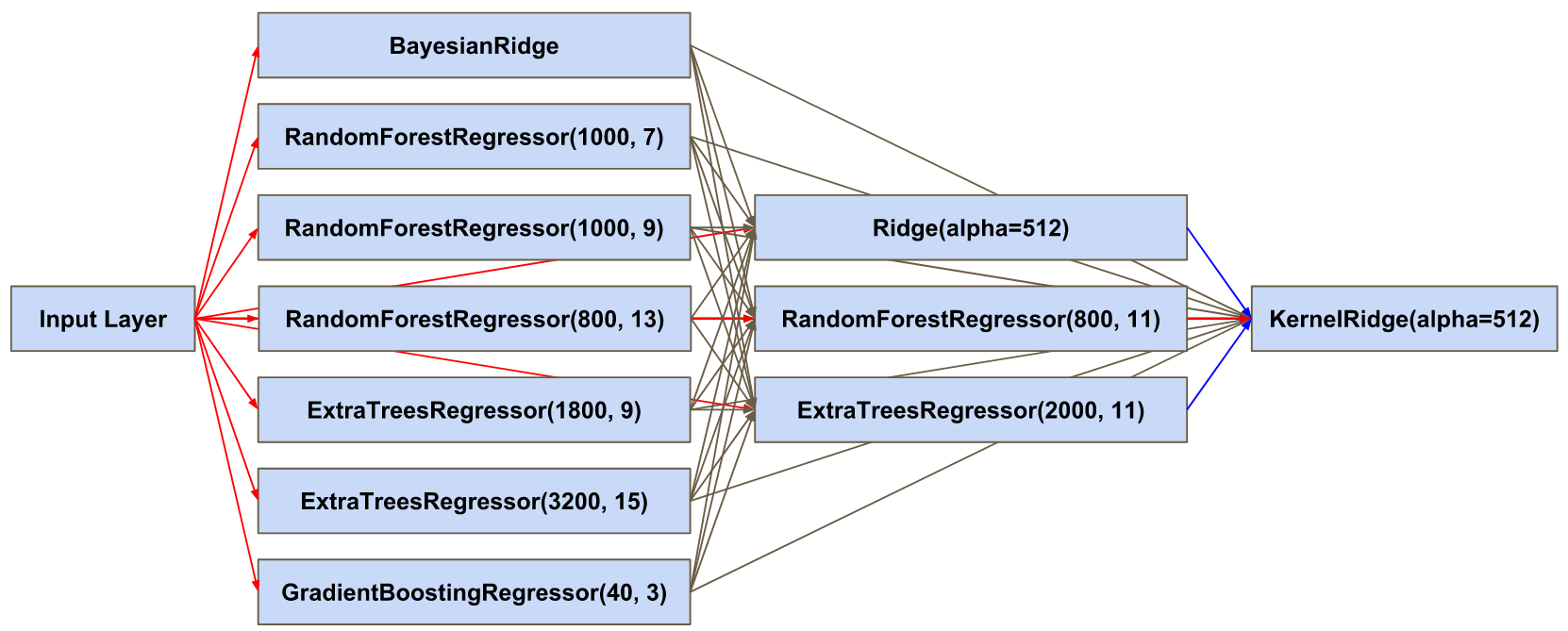}
    \caption{The architecture of proposed StackNet. For the ensemble-based regressor, the number of trees and the maximum depth of each tree are indicated in the first and second number, respectively.}
    \label{fig:stacknet}
\end{figure}

We adapt the StackNet architecture for our problem based on the following ideas: (i) including more models which have similar prediction performance, (ii) having a linear model in each layer (iii) placing models with better performance on a higher layer, and (iv) increasing the diversity in each layer. The resulting StackNet, shown in Fig.~\ref{fig:stacknet}, consists of three layers and 11 models. These models include one Bayesian ridge regressor \cite{mackay1992bayesian}, four random forest regressors \cite{breiman2001random}, three extra-trees regressors \cite{geurts2006extremely}, one gradient boosting regressor \cite{friedman2001greedy}, one kernel ridge regressor \cite{murphy2012machine}, and one ridge regressor.  The first layer has one linear regressor and five ensemble-based regressors, the second layer contains one linear regressor and two ensemble-based regressors, and the third layer only has one linear regressor. Each layer uses the predictions from all previous layers including the input layer. 

\subsection{Predicting Gf using Structural MR Images and StackNet}

\begin{figure}[htbp]
    \centering
    \includegraphics[width=\linewidth]{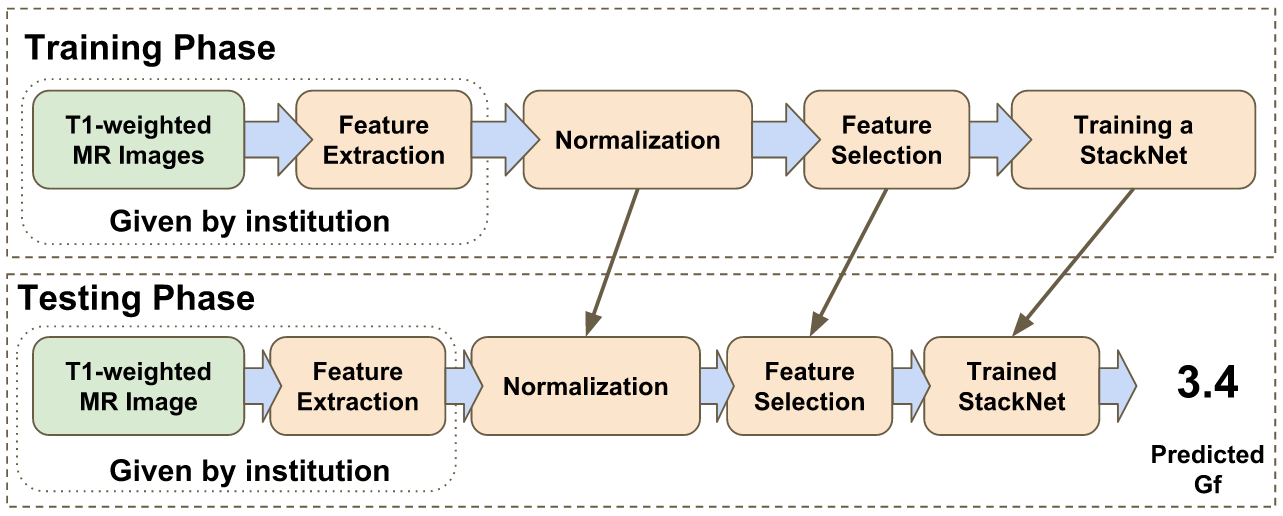}
    \caption{The framework of predicting fluid intelligence using MR-T1 images and a StackNet.}
    \label{fig:workflow}
\end{figure}

Fig.~\ref{fig:workflow} shows the framework of predicting the fluid intelligence scores using MR-T1 images and a StackNet. The framework is implemented with the scikit-learn \cite{sklearn_api,scikit-learn} Python library. In the training phase, features are extracted from the MR-T1 images of training and validation subjects. We then apply normalization and feature selection on the extracted features. In the end, these pre-processed features are used to train the StackNet in Fig.~\ref{fig:stacknet}. In the testing phase, features are extracted from the MR-T1 images of testing subjects, and the same feature pre-processing factors are applied to these extracted features. Thereafter, the pre-processed features are used with the trained StackNet to predict the fluid intelligence of the testing subjects. Details of each step are described below. 

The ABCD-NP-Challenge 2019 data includes pre-computed 122-dimension feature that characterizes the volume of brain tissues, i.e., gray matter, white matter, and cerebrospinal fluid, parcellated into SRI-24 \cite{rohlfing2010sri24} regions. The feature extracted for each subject is defined as $f_{i}(j)$ where $i$ is the index of subject and $j\in \{1,2,...,122\}$ is the index of feature dimension. 


\subsubsection{Normalization:} We apply a standard score normalization on each feature dimension, $\overline{f}_{i}(j) = (f_{i}(j) - \mu(j))/\sigma(j)$ where $i$ is the index of subject, $j$ is the index of feature dimension, and $\overline{f}_{i}(j)$ and $f_{i}(j)$ are the normalized and raw feature dimension $j$ of subject $i$, respectively. $\mu(j)$ and $\sigma(j)$ are the mean and standard deviation of feature dimension $j$, respectively. 

\subsubsection{Feature Selection:} Feature selection consists of three steps: (i) reducing the noise of data and generating an accurate representation of data through principal component analysis (PCA) \cite{tipping1999probabilistic} with maximum-likelihood estimator \cite{minka2001automatic} (ii) removing the feature dimensions with the low variance between subjects, and (iii) selecting 24 feature dimensions with the highest correlations to the ground-truth Gf scores through univariate linear regression tests. Thereafter, the feature dimensions shrink from 120 to 24.

\subsubsection{Training a StackNet:} Because the mean of the pre-residualized fluid intelligence for the training dataset ($\mu=0.05, \sigma=9.26$) and validation dataset ($\mu=-0.5, \sigma=8.46$) are quite different, we combine these two datasets ($\mu=0, \sigma=9.19$) for hyperparameter optimization and training a StackNet. 

\subsubsection{Predicting Fluid Intelligence:} In the testing phase, we first apply the same pre-processing factors used in the training phase to the extracted features of testing subjects. We then use the trained StackNet with these pre-processed features to predict the fluid intelligence scores of testing subjects. 

\subsubsection{Evaluation Metric:} The mean squared error (MSE) is used to calculate the error between the predicted Gf scores and the corresponding ground-truth Gf scores.

\subsection{Computing Feature Importance} \label{sec:importance}

We would like to discover the correlation between the Gf score and the brain tissue volume in a region. Thus, we compute the importance of each feature dimension, and higher importance represents a higher correlation. However, after feature selection, the original data space of dimension 122 is projected and reduced to a new space of dimension 24. In this new space, we first compute the importance of each feature dimension and then backward propagate it to the original data space of dimension 122. The details are explained as follows.

After dimensionality reduction, we obtain the individual correlations between the remaining 24 feature dimensions and the ground-truth Gf scores. These correlations are first converted to F values and then normalized w.r.t. the total F values of feature dimensions, i.e.,  $\overline{F}_{1} + \overline{F}_{2} + ... + \overline{F}_{24} = 1$, where $\overline{F}_{k}$ is the normalized F value of feature dimension $k\in \{1,2,...,24\}$. These normalized F values are used to build a normalized F vector as $\bm{\overline{F}}_{1\times24} = [\overline{F}_{1}, \overline{F}_{2}, ..., \overline{F}_{24}]$. We then use the corresponding eigenvectors and eigenvalues from the PCA transformation to build two matrices, $ \bm{U}_{122\times24} = [\vv{u}_{1}, \vv{u}_{2}, ..., \vv{u}_{24}]$ and $\bm{\Lambda}_{1\times24} = [\lambda_{1}, \lambda_{2}, ..., \lambda_{24}]$, where $\vv{u}_{k}$ and $\lambda_{k}$ are the corresponding eigenvector and eigenvalue for $\overline{F}_{k}$, respectively. The dimension of $\vv{u}_{k}$ is 122. We also normalize the eigenvalue vector w.r.t. the total value of eigenvalues, i.e.,  $\bm{\overline{\Lambda}}_{1\times24} = \bm{\Lambda}_{1\times24}/\lambda_{t}$, where $\lambda_{t} = \lambda_{1} + \lambda_{2} +...+ \lambda_{24}$. 
The normalization for eigenvalues is required to ensure that they have the same scale as the F values. 
Thereafter, we use $\bm{\overline{F}}, \bm{\overline{\Lambda}}$ and $\bm{U}$ to build the feature importance matrix $\bm{I}_{122\times24} = [\overline{F}_{1}\overline{\lambda}_{1}\vv{u}_{1}, \overline{F}_{2}\overline{\lambda}_{2}\vv{u}_{2}, ..., 
\overline{F}_{24}\overline{\lambda}_{24}\vv{u}_{24}]$. In the end, we sum up the absolute value of each element in every row of the matrix $\bm{I}_{122\times24}$,
$$ \vv{I}_{122\times1} = \sum_{n=1}^{24} |\overline{F}_{n}\overline{\lambda}_{n}\vv{u}_{n}|$$
$\vv{I}_{122\times1}$ is the feature importance vector in the original data space, and we also normalize it w.r.t. its total importance and rescale it,
$$\vv{I}_{nrm} = 100 \cdot \vv{I}_{122\times1} / \sum_{m=1}^{122}\vv{I}_{m}$$
Now, $\vv{I}_{nrm}$ is the normalized feature importance vector in the original data space of dimension 122, and each value of this vector represents the importance of a brain tissue volume in a region for the task of predicting the Gf scores. Higher importance represents higher correlation to the Gf score.

\section{Results and Discussion}

We examine the Gf prediction performance of individual models and StackNet on the combined dataset with 10-fold cross-validation, with the quantitative results shown in Table~\ref{tab:models}. The baseline is calculated by assigning the mean fluid intelligence ($\mu=0$) to every subject in the combined dataset. From Table~\ref{tab:models}, the performance of each model is better than the baseline of guessing the mean every subject, and the performance of the StackNet is better than every individual model within itself because it takes advantage of stacked generalization.
The proposed StackNet achieves a MSE of 94.2525 on the testing data as reported in the final leader board.

\begin{table}[htbp!]
    \centering
    \caption{The quantitative results of 11 models and StackNet with 10-fold cross-validation on the combined dataset. The bold number highlights the best performance. }
    \begin{tabular}{ l c }\hline
        Model & MSE \\\hline
        Baseline & 84.50 \\
        BayesianRidge & 82.62 \\
        Ridge(alpha=512) & 82.61 \\
        KernelRidge(alpha=512) & 82.61\\
        GradientBoostingRegressor(n\_estimators=40, max\_depth=3) & 83.60 \\
        RandomForestRegressor(n\_estimators=1000, max\_depth=7) & 83.07 \\
        RandomForestRegressor(n\_estimators=1000, max\_depth=9) & 83.09 \\
        RandomForestRegressor(n\_estimators=800, max\_depth=11) & 83.07 \\
        RandomForestRegressor(n\_estimators=800, max\_depth=13) & 83.11 \\
        ExtraTreesRegressor(n\_estimators=1800, max\_depth=9) & 83.16 \\
        ExtraTreesRegressor(n\_estimators=2200, max\_depth=11) & 83.10 \\
        ExtraTreesRegressor(n\_estimators=3200, max\_depth=15) & 83.16 \\ 
        StackNet & \textbf{82.42} \\ \hline
    \end{tabular}
    
    \label{tab:models}
\end{table}

The proposed StackNet in Fig.~\ref{fig:stacknet} is different from the StackNet which is used to report the MSE on the validation leader board. 
The StackNet used to report the MSE on the validation leader board has two layers and 8 models, and it achieves an MSE of 84.04 and 70.56 (rank 7 out of 17 teams) on the training and validation set, respectively. 
However, we noticed that statistics between the training set and validation are quite different, so we decided to combine these two datasets and work on this combined dataset ($\mu=0$ and $\sigma=9.19$) using 10-fold cross-validation. 
In addition, we also ensured that the mean and standard deviation of each fold is similar to the mean and standard deviation of combined dataset.

Second, we compute the importance of each dimension of the extracted feature by leveraging the F score from feature selection and eigenvectors and eigenvalues from PCA as described in Section~\ref{sec:importance}. Each dimension of the extracted feature corresponds to the volume of a certain type of brain tissue in a certain region. Table~\ref{tab:feature_importance_top10} and Table~\ref{tab:feature_importance_least10} show the top 10 most and least important feature dimensions for the task of predicting Gf, respectively, and higher importance represents higher correlation to the Gf scores. 

In conclusion, we demonstrate that the proposed StackNet with the distribution of different brain tissues in different brain parcellation regions has the potential to predict fluid intelligence in adolescents.

\begin{table}[h]
    \centering
    \caption{The top 10 most important variables for the task of predicting Gf}
    \begin{tabular}{c c}\hline
        Variable description & Importance \\\hline
        Pons white matter volume & 1.18 \\
        Right insula gray matter volume & 1.13 \\
        Right inferior temporal gyrus gray matter volume & 1.11 \\
        Corpus callosum white matter volume & 1.08 \\
        Cerebellum hemisphere white matter right volume & 1.07 \\
        Cerebellum hemisphere white matter left volume & 1.06 \\
        Left inferior temporal gyrus gray matter volume	& 1.06 \\
        Left insula gray matter volume & 1.06 \\
        Left superior frontal gyrus, orbital part gray matter volume & 1.05 \\
        Left opercular part of inferior frontal gyrus gray matter volume & 1.05 \\
        
    \hline
    \end{tabular}
    \label{tab:feature_importance_top10}
\end{table}

\begin{table}[h]
    \centering
    \caption{The top 10 least important variables for the task of predicting Gf}
    \begin{tabular}{c c}\hline
        Variable description & Importance \\\hline
        Right hippocampus gray matter volume & 0.53 \\
        Right amygdala gray matter volume & 0.54 \\
        Left hippocampus gray matter volume & 0.56 \\
        Right caudate nucleus gray matter volume & 0.58 \\
        Right lobule IX of cerebellar hemisphere volume & 0.60 \\
        Right lobule X of cerebellar hemisphere (flocculus) volume & 0.60 \\ 
        Left lobule X of cerebellar hemisphere (flocculus) volume & 0.61 \\ 
        Right superior parietal lobule gray matter volume & 0.61 \\
        Left middle temporal pole gray matter volume & 0.63 \\
        Left lobule IX of cerebellar hemisphere volume & 0.63 \\
        
    \hline
    \end{tabular}
    \label{tab:feature_importance_least10}
\end{table}

\section*{Acknowledgement}
This research was partially supported by a National Institutes of Health (NIH) award \# 5R01NS103774-02.

\bibliographystyle{splncs04}
\bibliography{reference}

\end{document}